# Faster Projected GAN: Towards Faster Few-Shot Image Generation

Chuang Wang[1], Zhengping Li[1], Yuwen Hao[2], Lijun Wang[1], Xiaoxue Li[2]

[1]Dept. of information, North China University of Technology, Peking 100000, China
[2]Medical Innovation Research Division of the Chinese PLA General Hospital, Beijing 100853, China

## Abstract

In order to solve the problems of long training time, large consumption of computing resources and huge parameter amount of GAN network in image generation, this paper proposes an improved GAN network model, which is named Faster Projected GAN, based on Projected GAN. The proposed network is mainly focuses on the improvement of generator of Projected GAN.  By introducing depth separable convolution (DSC), the number of parameters of the Projected GAN is reduced, the training speed is accelerated, and memory is saved. Experimental results show that on ffhq-1k, art-painting, Landscape and other few-shot image datasets, a 20% speed increase and a 15% memory saving are achieved. At the same time, FID loss is less or no loss, and the amount of model parameters is better controlled.   At the same time, significant training speed improvement has been achieved in the small sample image generation task of special scenes such as earthquake scenes with few public datasets.

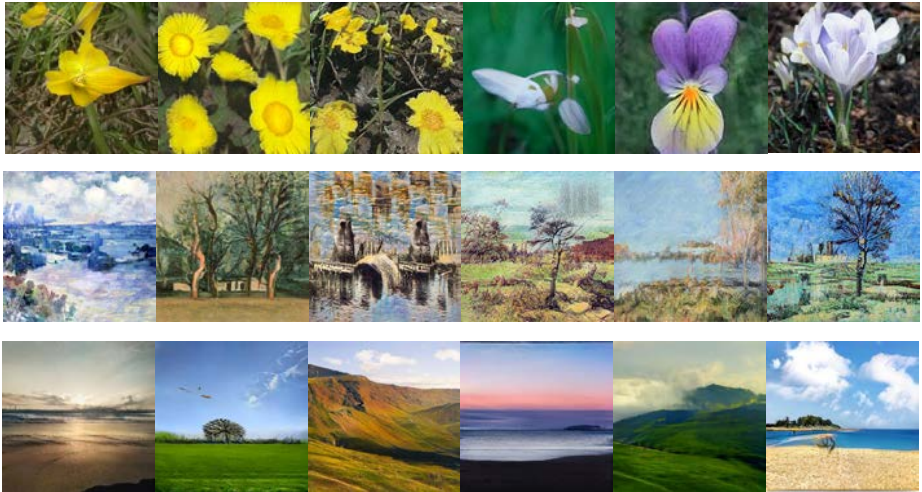

Figure 1: The generation results of our Faster Projected GAN. From top to bottom, the images are trained and generated on the flowers17, art-painting, and landscape $256^2$ resolution datasets.

## 1.  Introduction

Images are of great significance to social life. A picture is worth a thousand words. Images are very important in information transmission. Especially for special fields such as aerospace, medical, and criminal investigation, high-quality images can often play an important role [1]. Generative Adversarial Networks (GAN) [3], which enables the generator to generate random noise into samples through adversarial learning. Due to its special and novel adversarial training ideas, GAN has attracted the attention of a large number of researchers. Many improved models have achieved good results and have become the mainstream choice for image generation.

Few-shot learning is an important research field of artificial intelligence. The few-shot learning model should be trained on limited samples and have the ability to generate new samples that have not appeared in the training process [2]. The scarcity of data is often an important problem faced in this field. The collection of some data is relatively difficult, and labeling massive data will also consume a lot of manpower and financial resources. For some data, it is impossible to obtain more samples. In these fields, few-shot learning is an urgent problem that needs to be solved, and it is also an important development direction of artificial intelligence. In the field of small sample image generation, our work mainly focuses on improving the Projected GAN model. By introducing depth separable convolution (DSC), we simplify the network structure and propose Faster Projected GAN, which accelerates the training of the model while ensuring FID. The loss is within an acceptable range and memory is saved. We conducted experiments on large datasets and small sample datasets respectively, and the results show that Faster Projected GAN performs better on small sample datasets. At the same time, we found that the effects of using DSC on Discriminator and Generator are different, and using DSC on Generator has better results.

## 2. Related Work

**GANs:** Generative adversarial networks (GANs) have made significant progress in deep learning since their introduction in [3]. Due to the learning method of GAN network adversarial training, it shows strong capabilities in data generation. GAN has been successfully used in various visual fields, including image generation [4][5], video generation [6], image to image Translation[7][8] etc. The GAN model consists of a generator and a discriminator, and the two networks are updated alternately in an adversarial manner. Training GAN is also difficult because of its adversarial training method, because it requires a large amount of data and computing resources. At the same time, the adversarial training method makes the training process unstable. If one party is too strong, it will cause the model to collapse. When given limited data, discriminators tend to overfit, resulting in poor quality generation. In order to prevent the discriminator from overfitting, researchers have proposed many works. Different data augmentation techniques, including differentiable [9], non-leakage [10] and adaptive pseudo-augmentation [11], have been developed to expand limited training data. For example, Lecam [12] regularizes the output of the discriminator to avoid overfitting. Projected GAN [13] is not based on traditional transfer learning, but uses the prior knowledge of the pre-trained model to gradually increase the level of the generator and discriminator to gradually increase the resolution of the generated image. Although Projected GAN is not based on the traditional transfer learning paradigm, it utilizes information from previous levels during the training process. This progressive approach is somewhat similar to the concept of transfer learning, adapting to more complex tasks through gradual learning. The work of this article is to simplify the hierarchical training process of Projected GAN, improve the training speed and reduce memory usage, while ensuring the quality of image generation.

**Few-shot Image Generation:** The purpose of few-shot image generation is to generate new and diverse image instances while preventing overfitting to a small number of training images. Existing small sample image generation methods can be roughly divided into optimization-based methods, fusion-based methods and transformation-based methods. Optimization-based methods [14-16] combine meta-learning and adversarial learning to generate images of unseen categories by fine-tuning the model. Fusion-based methods fuse local features [17] or input images [18][19] to synthesize new images. GMN [20] combines VAE [21] with a matching network [22] to capture few-sample distributions. Matching GAN [19] matches random vectors with given real images and maps the fused features to new images. F2GAN [18] further improves Matching GAN through fusion and filling paradigms. LoFGAN [17] improves the generation quality by fusing local representations with semantic similarity. Transformation-based methods [23-24] capture cross-category or intra-category transformations to generate new data of unseen categories. Optimization-based methods suffer from overfitting and overly complex design of the meta-learning process. Transform-based methods suffer from generalization limitations and may perform poorly on certain datasets, while there are some challenges in deciding which network layers to transform when designing the model. Our model, Faster Projected GAN, is improved on Projected GAN. It is a fusion-based method that uses the prior knowledge of the pre-trained model to train GAN to achieve small sample image generation in specific categories.

**Depthwise Separable Convolution:** While the use of spatially separable convolutions in neural networks has a long history, dating back to at least 2012 [25] (but probably earlier), depthwise separable convolutions are somewhat newer. Laurent Sifre developed depthwise separable convolution during his internship at Google Brain in 2013 and used it in AlexNet, achieving a small improvement in accuracy and a large increase in convergence speed, and a significant reduction in model size. An overview of his work was first made public in a talk at ICLR 2014 [26]. Sifre's paper [27] reports detailed experimental results. This preliminary work on depthwise separable convolutions was inspired by previous work on transformation-invariant scattering by Sifre and Mallat [27, 28]. Later, depthwise separable convolutions were used as the first layer in Inception V1 and Inception V2 [29, 30]. Within Google, Andrew Howard [31] introduced efficient mobile models called MobileNets using depthwise separable convolutions. Jin et al. 2014 [32] and Wang et al. 2016 [33] also did related work, aiming to use separable convolutions to reduce the size and computational cost of convolutional neural networks. In 2017 [34] used depthwise separable convolution to replace the ordinary convolution in Inception V3. The results proved that with the same number of parameters, the Xception model achieved better results due to the efficient use of parameters. In 2019 [35] proposed DepthwiseGANs, which used depthwise separable convolution to accelerate model training and showed better performance than StarGAN. At the same time, their experiments showed that using depthwise separable convolution on Generator has the best effect. In 2020 [36] changed the convolution order of Pointwise and Depthwise in depthwise separable convolution, achieving great improvements in MobileNet, further revealing that DSC implicitly relies on cross-core correlation. In the proposed network, DSC is intruded in Faster Projected GAN to improve the performance.

## 3. Network structure of Faster Projected GAN

Our improvement of the network structure of projected-gan mainly focuses on the improvement of Generator. Our experimental results are the same as [35] Ngxande's conclusion.

Using depth-separable convolution on Generator can achieve the best results. On Discriminator we We also tried to use depthwise separable convolution optimization. However, the results show that using depthwise separable convolution on Discriminator cannot speed up training while maintaining the FID effect. In fact, after using DSC on Discriminator to improve it, the FID effect becomes Very poor, so that the accelerated training is not worth the gain. Therefore, we did not change the network structure of Discriminator and used DSC optimization on Generator. However, the original convolution is maintained in the Skip-layer excitation module. We do not simplify the convolution in this module to prevent the model capacity from being too small and unable to exert the performance of this special module. The improved Generator network structure is shown in Figure 2.

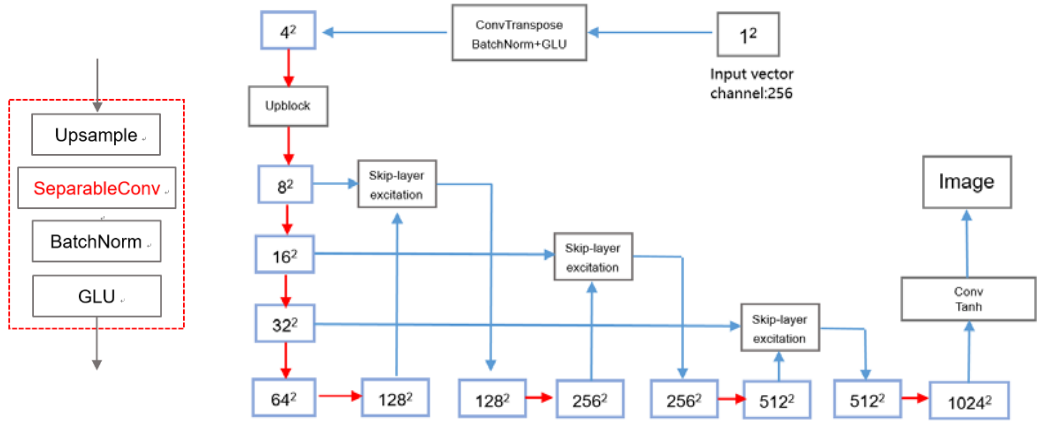

Figure 2: Network structure of Generator of Faster Projected GAN. The left is the improved upblock in our model which is represented as red arrows in the right figure. The right is the structure of the Generator.

## 4. Ablation

In order to verify the role of depth separable convolution (DSC) in Discriminator and Generator, we designed a comparative experiment. Faster pg DG uses the DSC module in both Discriminator and Generator, including SEblock which also uses DSC. Faster pg G only uses the DSC module in the Generator, excluding the SEblock module. We conducted ablation experiments on the self-built seismic , art-painting , landscape and animalface-cat dataset at $256^2$ scale, and the results are shown in the Figure 3.

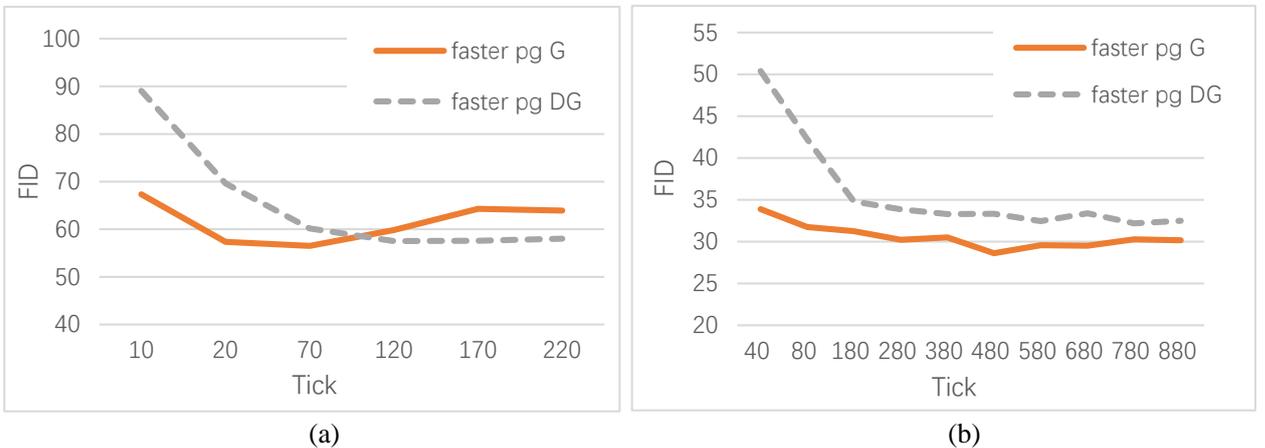

(a)          (b)

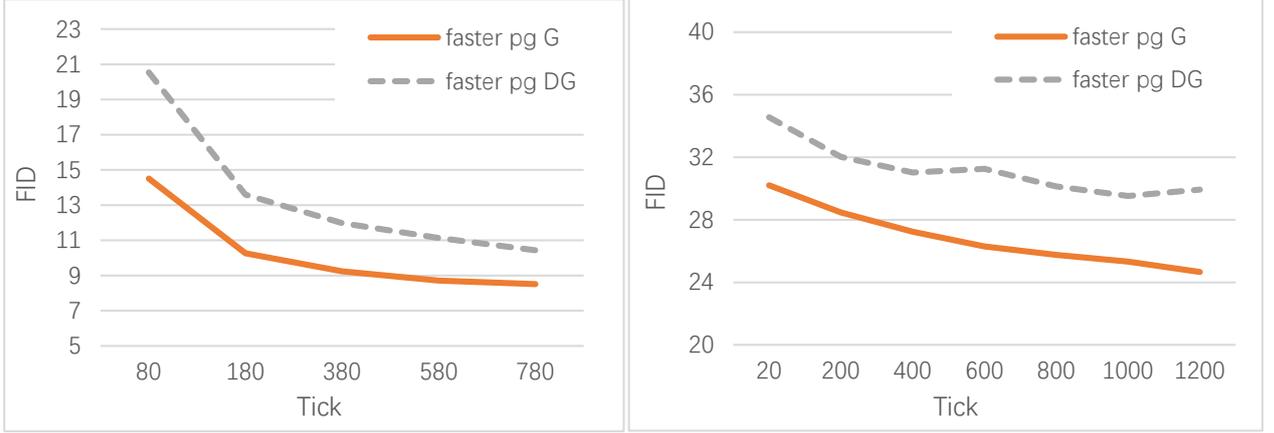

(c)　　　　　　　　　　　　　　　　(d)

Figure 3: Ablation experiment results. (a): Training process at $512^2$ resolution of the self-built seismic dataset. (b): Training process on art-painting $256^2$. (c): training process on landscape $256^2$. (d): Training process on animalface-cat $256^2$.

Table 1: Comparison of ablation experiment results on FID

|  | seismic ($512^{2)}$) | landscape ($256^2$) | artpainting ($256^{2)}$) | animalface-cat ($256^{2)}$) |
|---|---|---|---|---|
| Faster pg DG | 55.26 | 9.92 | 33.32 | 32.83 |
| Faster pg G | **53.59** | **7.68** | **28.60** | **27.69** |

The results of the ablation experiment show that the effects of using the DSC module on G and D are different. Using the DSC module on the D tends to have worse results than using the DSC module alone on the G. The reason is mainly related to the roles and functions of the two in the network, as well as the characteristics of depthwise separable convolution itself. Generators need to create complex and high-quality images, which usually involves generating high-dimensional image data from lower-dimensional noise vectors, a complex upsampling process. Since the task of the generator itself is complex and computationally intensive, using depthwise separable convolution can significantly reduce the computational burden and model parameters and increase the generation speed, which is a great advantage for the generator. The task of the discriminator is to judge whether the input image is real or generated by the generator, which is usually a relatively straightforward classification problem. Although using depthwise separable convolutions in the discriminator can also improve efficiency, the task of the discriminator is relatively simple and the need to improve computational efficiency may not be as pressing as that of the generator. For generators, reducing model capacity could help to prevent overfitting, train faster, and generate high-quality images. For the discriminator, an oversimplified model may reduce its ability to accurately judge authenticity. Especially when facing complex and diverse real images, accurate judgment is more important.

## 5. Experimental analysis and evaluation

In order to verify the effectiveness of the improved model, the advantages of Faster Projected GAN over the state-of-the-art model are demonstrated. Our experiments mainly compare indicators such as convergence speed and FID on small benchmark datasets. Tests were conducted on ffhq-1k, landscape, art-painting, pokemon, Obama and other datasets. We mainly conduct experiments on small sample datasets and images with a resolution of $256^2$. The experimental data on the FFHQ dataset shows that the improvement effect is not obvious on the large sample dataset. However, considering the speed advantage, compared with other models, the effect is not obvious. Unless otherwise stated, all experiments were completed on the single A40 GPU.

Table 2 compares the model FID index and training time between Faster Projected GAN and Projected GAN on the ffhq-1k, Landscape, Art-painting, and Pokemon datasets at a resolution of $256^2$. The results show that on the ffhq-1k resolution $256^2$ dataset, the FID loss is 2.05%, but the training time is reduced by 18.7%. On the Landscape dataset, the FID value is slightly inferior, with a loss of 0.26%, while the training time increases by 19.24%. On the Art-Painting dataset, the speed increased by 26.16% with a FID loss of 1.72%. On the Pokemon dataset, FID lost 11.01% and the speed increased by 23.6%. Finally, training is done on a self-built seismic scene small sample dataset. On $512^2$ resolution pictures, the dual improvement of FID and speed is achieved, and the effect is remarkable.

Compared with the Projected GAN model, Faster Projected GAN has obvious improvement effects. It achieves speed improvement and memory saving on small sample datasets. At the same time, the FID training effect is less reduced or even improved. This illustrates the huge potential and application value of depth-separable convolution in GAN network structure optimization.

Table 2: Improvement effect of Faster Projected GAN

|  | FID | Time (h) | FID | Time (h) | FID | Time (h) | FID | Time (h) |
|---|---|---|---|---|---|---|---|---|
|  | FFHQ-1k ($256^2$) | | Landscape ($256^2$) | | Art Painting ($256^2$) | | Pokemon ($256^2$) | |
| Projected GAN | 23.93 | 26.80 | 7.66 | 42.1 | 28.12 | 27.93 | 26.42 | 55.05 |
| Faster Projected GAN | 24.42 | 21.79 | 7.68 | 34.0 | 28.60 | 20.63 | 29.33 | 42.06 |
| Improve Percentage | -2.05% | 18.70% | -0.26% | 19.24% | -1.72% | 26.16% | -11.01% | 23.60% |

Table 3: Comparison of FID results on the $256^2$ small sample dataset

| | | | AnimalFace-Dog | AnimalFace-Cat | Obama | Panda | Grumpy-cat |
|---|---|---|---|---|---|---|---|
| | Image number | | 389 | 160 | 100 | 100 | 100 |
| Training time on one RTX 2080-Ti | 20 hours | StyleGAN2 | 58.85 | 42.44 | 46.87 | 12.06 | 27.08 |
| | 5 hours | FastGAN | 50.66 | 35.11 | 41.05 | 10.03 | 26.65 |
| | 2 hours | Faster Projected GAN | **38.13** | **28.66** | **30.81** | **8.25** | **23.28** |

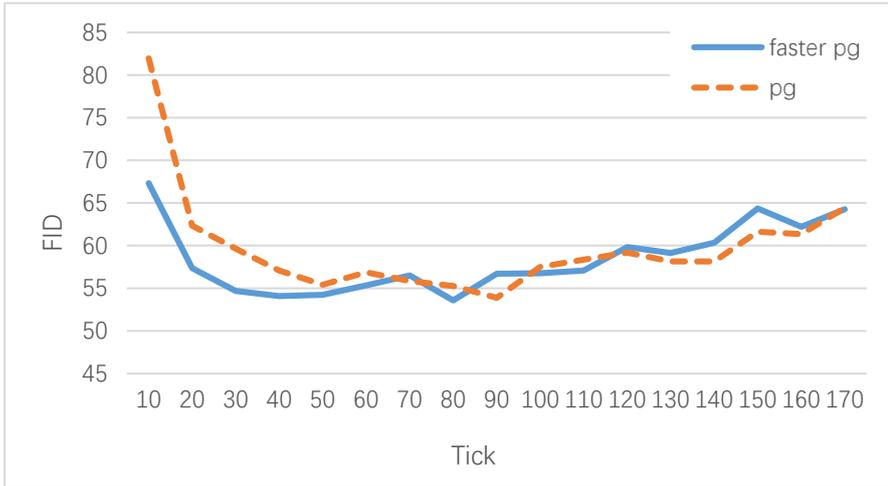

Figure 4: Training results on the self-built seismic $512^2$ dataset

Table 4: Detailed Metrics on Small Datasets ($256^2$)

|  | Small Datasets($256^2$) | | | |
|---|---|---|---|---|
|  | Art Painting | Landscape | Flowers | Pokemon |
|  | FID ↓ | | | |
| StyleGAN-ADA | 43.07 | 15.99 | 21.66 | 40.38 |
| FastGAN | 44.02 | 16.44 | 26.23 | 81.86 |
| Projected GAN | **27.96** | **6.92** | **13.86** | **26.41** |
| Fast Projected GAN | 28.60 | 8.05 | 14.67 | 29.09 |
|  | KID x $10^3$ ↓ | | | |
| StyleGAN-ADA | 10.23 | 4.39 | 3.56 | 13.49 |
| FastGAN | 13.00 | 3.40 | 6.61 | 80.30 |
| Projected GAN | **1.25** | 1.30 | **0.38** | **1.32** |
| Fast Projected GAN | 1.48 | **0.57** | 0.53 | 3.22 |
|  | Precision ↑ | | | |
| StyleGAN-ADA | 0.691 | 0.709 | 0.731 | 0.735 |
| FastGAN | **0.858** | 0.768 | 0.611 | 0.731 |
| Projected GAN | 0.762 | **0.774** | **0.816** | **0.809** |
| Fast Projected GAN | 0.696 | 0.742 | 0.743 | 0.797 |
|  | Recall ↑ | | | |
| StyleGAN-ADA | 0.218 | 0.213 | 0.095 | 0.197 |
| FastGAN | 0.044 | 0.160 | 0.100 | 0.004 |
| Projected GAN | 0.239 | **0.258** | 0.058 | 0.259 |
| Fast Projected GAN | **0.351** | 0.248 | **0.135** | **0.286** |

# 6. Conclusion

To address the problem of small sample generation, this paper proposes an improved small sample adversarial generation network, Faster Projected GAN, by combining depth-separable convolution and Projected GAN from the perspective of improving training speed and reducing memory. DSC essentially decomposes the original convolution kernel to achieve the purpose of reducing the number of parameters. Splitting the convolution kernel essentially reduces the network capacity and increases the depth of the network, which is conductive to the network extracting deep features and is suitable for samples with fewer features. Comparison experiments were conducted on public small sample datasets and self-built datasets. The results show that the Faster Projected GAN can accelerate training and save video memory. At the same time, the FID loss of generated images is less, and the training speed is significantly improved. At the same time, due to the different structures of Generator and Discriminator in the GAN network, the effect is significantly different after using the DSC module, and the effect of using DSC on the Generator is better.

# 7. References

[1] Akash Tayal,Jivansha Gupta,Arun Solanki,et al.Correction to: dl-cnn-based approach with image processing techniques for diagnosis of retinal diseases[J].Multimedia Systems,2021(prepublish):1-1.
[2] Wang, Yaqing, et al. "Generalizing from a few examples: A survey on few-shot learning." *ACM computing surveys (csur)* 53.3 (2020): 1-34.
[3] Goodfellow, I., Pouget-Abadie, J., Mirza, M., Xu, B., Warde-Farley, D., Ozair, S., Courville, A., Bengio, Y.: Generative adversarial nets. Advances in neural infor- mation processing systems 27 (2014)
[4] Karras, T., Laine, S., Aittala, M., Hellsten, J., Lehtinen, J., Aila, T.: Analyzing and improving the image quality of stylegan. In: Proceedings of the IEEE/CVF conference on computer vision and pattern recognition. pp. 8110–8119 (2020)
[5] Karras, T., Aittala, M., Laine, S., Härkönen, E., Hellsten, J., Lehtinen, J., Aila, T.: Alias-free generative adversarial networks. Advances in Neural Information Processing Systems 34 (2021)
[6] Wang, L., Ho, Y.S., Yoon, K.J., et al.: Event-based high dynamic range image and very high frame rate video generation using conditional generative adversarial networks. In: Proceedings of the IEEE/CVF Conference on Computer Vision and Pattern Recognition. pp. 10081–10090 (2019)
[7] Park, T., Efros, A.A., Zhang, R., Zhu, J.Y.: Contrastive learning for unpaired image-to-image translation. In: European Conference on Computer Vision. pp. 319–345. Springer (2020)
[8] Richardson, E., Alaluf, Y., Patashnik, O., Nitzan, Y., Azar, Y., Shapiro, S., Cohen- Or, D.: Encoding in style: a stylegan encoder for image-to-image translation. In: Proceedings of the IEEE/CVF Conference on Computer Vision and Pattern Recog- nition. pp. 2287–2296 (2021)
[9] Zhao, S., Liu, Z., Lin, J., Zhu, J.Y., Han, S.: Differentiable augmentation for data- efficient gan training. Advances in Neural Information Processing Systems 33, 7559–7570 (2020)
[10] Karras, T., Aittala, M., Hellsten, J., Laine, S., Lehtinen, J., Aila, T.: Training generative adversarial networks with limited data. Advances in Neural Information Processing Systems 33, 12104–12114 (2020)
[11] Jiang, L., Dai, B., Wu, W., Loy, C.C.: Deceive d: Adaptive pseudo augmentation for gan training with limited data. Advances in Neural Information Processing Systems 34 (2021)
[12] Tseng, H.Y., Jiang, L., Liu, C., Yang, M.H., Yang, W.: Regularizing generative adversarial networks under limited data. In: Proceedings of the IEEE/CVF Con- ference on Computer Vision and Pattern Recognition. pp. 7921–7931 (2021)
[13] Sauer, Axel, et al. "Projected gans converge faster." Advances in Neural Information Processing Systems 34 (2021): 17480-17492.
[14] Louis Clouâtre and Marc Demers. FIGR: few-shot image generation with reptile. arXiv preprint arXiv:1901.02199, 2019. 1, 2, 5
[15] Weixin Liang, Zixuan Liu, and Can Liu. Dawson: A domain adaptive few shot generation framework. arXiv preprint arXiv:2001.00576, 2020. 1, 2, 5
[16] Aniwat Phaphuangwittayakul, Yi Guo, and Fangli Ying. Fast adaptive meta-learning for few-shot image generation. IEEE Transactions on Multimedia, pages 1–1, 2021. 2
[17] Gu, Z., Li, W., Huo, J., Wang, L., Gao, Y.: Lofgan: Fusing local representations for few-shot image generation. In: Proceedings of the IEEE/CVF International Conference on Computer Vision. pp. 8463–8471 (2021)

[18] Hong, Y., Niu, L., Zhang, J., Zhao, W., Fu, C., Zhang, L.: F2gan: Fusing-and-filling gan for few-shot image generation. In: Proceedings of the 28th ACM International Conference on Multimedia. pp. 2535–2543 (2020)

[19] Hong, Y., Niu, L., Zhang, J., Zhang, L.: Matchinggan: Matching-based few-shot image generation. In: 2020 IEEE International Conference on Multimedia and Expo (ICME). pp. 1–6. IEEE (2020)

[20] Bartunov, S., Vetrov, D.: Few-shot generative modelling with generative matching networks. In: International Conference on Artificial Intelligence and Statistics. pp. 670–678. PMLR (2018)

[21] Kingma, D.P., Welling, M.: Auto-encoding variational bayes. arXiv preprint arXiv:1312.6114 (2013)

[22] Vinyals, O., Blundell, C., Lillicrap, T., Kavukcuoglu, K., Wierstra, D.: Matching networks for one shot learning. In: Proceedings of the 30th International Conference on Neural Information Processing Systems. pp. 3637–3645 (2016)

[23] Antreas Antoniou, Amos Storkey, and Harrison Edwards. Data augmentation generative adversarial networks. arXiv preprint arXiv:1711.04340, 2017. 1, 2, 5, 6

[24] Y. Hong, Li Niu, Jianfu Zhang, Jing Liang, and Liqing Zhang. Deltagan: Towards diverse few-shot image generation with sample-specific delta. In CVPR, 2020. 1, 2, 5, 6, 7

[25] F. Mamalet and C. Garcia. Simplifying ConvNets for Fast Learning. In International Conference on Artificial Neural Networks (ICANN 2012), pages 58–65. Springer, 2012.

[26] V. Vanhoucke. Learning visual representations at scale. ICLR, 2014.

[27] L. Sifre. Rigid-motion scattering for image classification, 2014. Ph.D. thesis.

[28] L. Sifre and S. Mallat. Rotation, scaling and deformation invariant scattering for texture discrimination. In 2013 IEEE Conference on Computer Vision and Pattern Recognition, Portland, OR, USA, June 23-28, 2013, pages 1233–1240, 2013.

[29] C. Szegedy, W. Liu, Y. Jia, P. Sermanet, S. Reed, D. Anguelov, D. Erhan, V. Vanhoucke, and A. Rabinovich. Going deeper with convolutions. In Proceedings of the IEEE Conference on Computer Vision and Pattern Recognition, pages 1–9, 2015.

[30] S. Ioffe and C. Szegedy. Batch normalization: Accelerating deep network training by reducing internal covariate shift. In Proceedings of The 32nd International Conference on Machine Learning, pages 448–456, 2015.

[31] A. Howard. Mobilenets: Efficient convolutional neural net- works for mobile vision applications. Forthcoming.

[32] J. Jin, A. Dundar, and E. Culurciello. Flattened convolutional neural networks for feedforward acceleration. arXiv preprint arXiv:1412.5474, 2014.

[33] M. Wang, B. Liu, and H. Foroosh. Factorized convolutional neural networks. arXiv preprint arXiv:1608.04337, 2016.

[34] Chollet F .Xception: Deep Learning with Depthwise Separable Convolutions[C]//2017 IEEE Conference on Computer Vision and Pattern Recognition (CVPR).IEEE, 2017.DOI:10.1109/CVPR.2017.195.

[35] M. Ngxande, J. -R. Tapamo and M. Burke, "DepthwiseGANs: Fast Training Generative Adversarial Networks for Realistic Image Synthesis," 2019 Southern African Universities Power Engineering Conference/Robotics and Mechatronics/Pattern Recognition Association of South Africa (SAUPEC/RobMech/PRASA), Bloemfontein, South Africa, 2019, pp. 111-116, doi: 10.1109/RoboMech.2019.8704766.

[36] Haase D , Amthor M .Rethinking Depthwise Separable Convolutions: How Intra-Kernel Correlations Lead to Improved MobileNets[J].IEEE, 2020.DOI:10.1109/CVPR42600.2020.01461.

[37] Paszke, Adam, et al. "Pytorch: An imperative style, high-performance deep learning library." *Advances in neural information processing systems* 32 (2019).